\documentclass[letterpaper,10pt,conference]{ieeeconf}

\providecommand{\IEEEoverridecommandlockouts}{}
\IEEEoverridecommandlockouts
\providecommand{\overrideIEEEmargins}{}
\usepackage{graphicx}
\usepackage{booktabs}
\usepackage[table]{xcolor}
\usepackage{array}
\usepackage{amsmath}
\usepackage{amssymb}
\usepackage{cuted}
\usepackage{capt-of}
\usepackage{multirow}
\usepackage[hyphens]{url}
\usepackage{iftex}
\ifPDFTeX
  \usepackage[T1]{fontenc}
  \usepackage{mathptmx}
\else
  \usepackage{fontspec}
  \defaultfontfeatures{Ligatures=TeX}
\fi

\definecolor{tableheader}{RGB}{224,232,240}
\definecolor{tablealt}{RGB}{249,251,253}
\definecolor{tableadapt}{RGB}{255,249,232}
\definecolor{tablecontrol}{RGB}{245,247,249}
\definecolor{tablegain}{RGB}{237,248,239}
\definecolor{tableours}{RGB}{231,243,235}
\definecolor{tablestrong}{RGB}{237,244,250}
\newcommand{\tablehead}[1]{\textcolor{black}{\textbf{#1}}}
\newcommand{\safeincludegraphics}[2][]{%
  \IfFileExists{#2}{%
    \includegraphics[#1]{#2}%
  }{%
    \fbox{\parbox[c][3.2cm][c]{0.94\linewidth}{%
      \centering\footnotesize Missing figure file:\\[0.4em]
      \texttt{\detokenize{#2}}%
    }}%
  }%
}

\title{\LARGE \bf
INHerit-SG: \underline{In}cremental \underline{H}i\underline{er}arch\underline{i}cal Seman\underline{t}ic \underline{S}cene \underline{G}raphs \\
for Complex Embodied Retrieval
}

\author{%
\authorblockN{%
YukTungSamuel Fang\textsuperscript{1},
Zhikang Shi\textsuperscript{1},
Jiabin Qiu\textsuperscript{1},
Zixuan Chen\textsuperscript{1},\\
Jieqi Shi\textsuperscript{1*},
Hao Xu\textsuperscript{2},
Jing Huo\textsuperscript{1*},
Yang Gao\textsuperscript{1}
}
\authorblockA{%
\textsuperscript{1}State Key Laboratory for Novel Software Technology, 
Nanjing University, Nanjing, China
}
\authorblockA{%
\textsuperscript{2}Beihang University, Beijing, China
}
\authorblockA{%
\textsuperscript{*}Corresponding authors:
\texttt{isjieqi@nju.edu.cn},
\texttt{huojing@nju.edu.cn}
}
\authorblockA{%
Other emails:
\texttt{\{231880023, 221900090, 221900358\}@smail.nju.edu.cn},\\
\texttt{\{chenzx, gaoy\}@nju.edu.cn},
\texttt{xuhao3e8@buaa.edu.cn}.
}
}

\begin{document}
\maketitle
\thispagestyle{empty}
\pagestyle{empty}

\begin{strip}
\centering
\safeincludegraphics[width=0.96\textwidth]{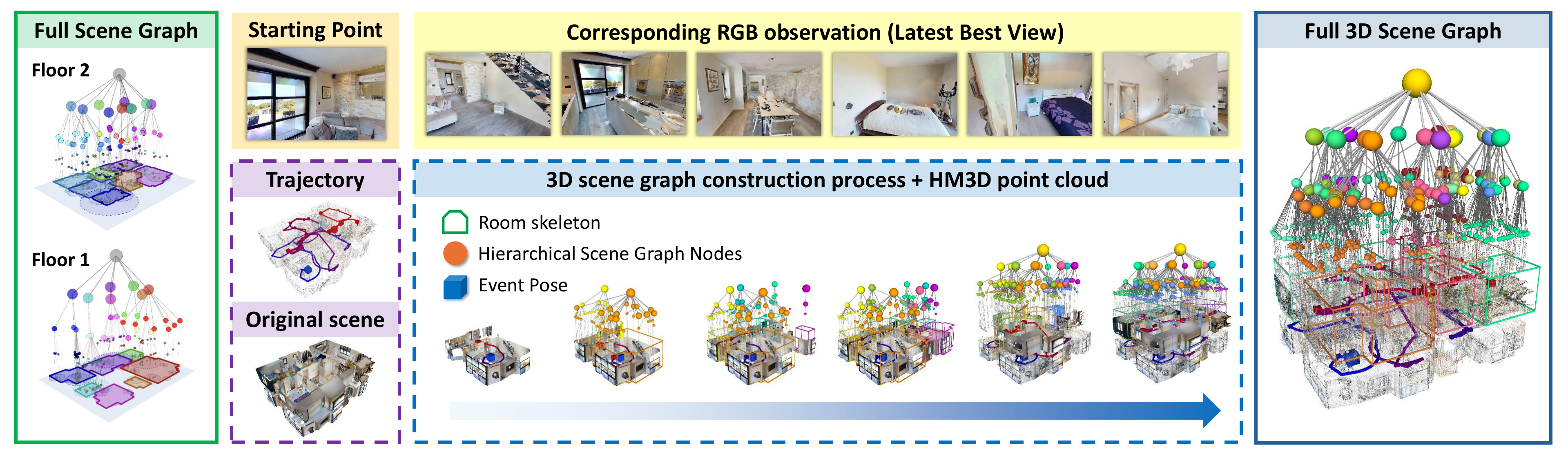}
\captionof{figure}{\textbf{INHerit-SG builds an incrementally maintained hierarchical scene graph for complex object retrieval.}}
\label{fig:teaser}
\end{strip}

\begin{abstract}
Object retrieval in 3D scenes remains difficult when queries combine hierarchical context, attributes, negation, and chained spatial relations. Existing semantic maps, scene graphs, and agentic memory support open-vocabulary retrieval, but jointly resolving these constraints remains challenging. In this work, we propose INHerit-SG, an incremental hierarchical semantic scene graph with an accompanying retrieval method for complex embodied retrieval. Asynchronous geometric and semantic processing and structural event-driven maintenance organize observations into a hierarchy whose node semantics and spatial relations are recorded as natural language text. The same memory supports direct reading and reasoning by more capable language models or query parsing followed by structured retrieval. Controlled comparisons show that direct reading is more accurate with powerful hosted models, while structured retrieval improves the accuracy of smaller local models and lowers per-query API cost for hosted models. We also construct HM3DSem-SQR, containing 6,084 queries from 36 HM3D-Sem scenes, and evaluate on Main-864, a class-balanced subset spanning all 36 scenes, and in three real-world environments. Experiments show that INHerit-SG outperforms the strongest baselines in both simulator and real-world evaluations. The largest gains come on attribute and compositional queries (7.6 and 6.3 strict Top-1 points on Main-864), with a 9.8-point overall success margin on real robots. Project Page: \url{https://fangyuktung.github.io/INHeritSG.github.io/}
\end{abstract}

\vspace{-10pt}

\section{Introduction}

Retrieving an object from a category-only instruction is relatively straightforward. In embodied settings, however, human instructions typically compose hierarchical scope, attributes, negation, and spatial relations to reference objects, and these jointly determine the target. A single global text--scene similarity score cannot adequately express such composed constraints.

Open-vocabulary mapping methods store semantics as dense features attached to points or objects \cite{gu2023conceptgraphs,peng2023openscene,huang2023vlmaps}, which supports flexible similarity matching but provides no explicit structure for composed constraints. Scene graphs, in contrast, explicitly organize instances into spatial hierarchies \cite{hughes2022hydra,werby2024hovsg,maggio2024clio,werby2026keysghierarchicalkeyframebased3d}. Their retrieval interfaces range from similarity matching to language-model parsing and hierarchical selection. However, incrementally constructing scene memory that preserves query logic, and jointly resolving attributes, negation, and dependent reference relations, remains challenging.

We study what spatial memory an embodied agent should construct to respond directly to complex natural language constraints, and argue that representation, construction, and retrieval must be designed together. Our answer is \textbf{INHerit-SG}, an incremental four-level semantic scene graph that organizes the environment into floors, rooms, functional areas, and objects while retaining object-level visual and linguistic evidence. All node semantics and spatial relations are recorded as natural language text. Language compositionally encodes attributes, negation, and reference relations \cite{fodor1988connectionism}. Dense vector representations obscure exactly this structure \cite{lake2018scan,kassner2020negated}. Construction separates geometric processing from slower foundation-model inference through asynchronous execution. At query time, language models can read the same memory directly, or access it through structured retrieval over typed constraints, preserving query logic rather than compressing it into a single embedding.

In summary, our contributions are threefold:
\begin{enumerate}
  \item We propose INHerit-SG, an incremental hierarchical semantic scene graph for compositional object retrieval. It integrates asynchronous geometric and semantic processing with event-driven hierarchy maintenance to organize RGB-D observations into a queryable spatial memory.

  \item We develop a lightweight structured retrieval method that combines typed query constraints, hierarchical and relational evidence, multi-path candidate retrieval, and selective visual reranking. The method enables compact language models to query scene memory without processing the complete graph.

  \item We introduce HM3DSem-SQR, a benchmark comprising 6,084 queries across 36 HM3D-Sem scenes, with a manually reviewed, class-balanced evaluation subset. Experiments on HM3DSem-SQR, ViGiL3D, and three real-world environments, covering retrieval accuracy, generalization, and query cost, demonstrate that INHerit-SG achieves state-of-the-art retrieval accuracy through a lightweight configuration.
\end{enumerate}

\section{Related Work}

\subsection{Open-Vocabulary 3D Representations}

Open-vocabulary mapping methods attach semantic features to points or objects, which enables flexible similarity-based retrieval \cite{gu2023conceptgraphs,peng2023openscene,huang2023vlmaps}. Hierarchical scene graphs additionally organize objects by spatial context \cite{hughes2022hydra,werby2024hovsg,maggio2024clio,werby2026keysghierarchicalkeyframebased3d}. Their retrieval interfaces range from similarity matching to language-model parsing and level-by-level selection. Jointly resolving attributes, negation, and dependent reference relations remains challenging, particularly when early selections restrict subsequent candidates. We address this through jointly designed incremental memory construction and query-conditioned retrieval.

\subsection{Hierarchical and Dynamic Scene Graph Construction}
\label{RW:B}
Hierarchical scene graphs organize object-centric memories into spatial and functional structures. Hydra models indoor spatial hierarchies, HOV-SG incorporates open-vocabulary semantics, and OpenFunGraph groups objects by function \cite{hughes2022hydra,werby2024hovsg,zhang2025openfungraph}. Open Scene Graphs support open-world object-goal navigation \cite{loo2025openscenegraphs}, while Khronos, DualMap, and dynamic open-vocabulary scene graphs address map maintenance and long-term mobile manipulation \cite{schmid2024khronos,jiang2025dualmap,10909193}. These advances improve spatial organization and map maintenance, but spatial hierarchy alone does not resolve how retrieval should combine multiple query constraints.

Our closest counterparts, HOV-SG and KeySG \cite{werby2024hovsg,werby2026keysghierarchicalkeyframebased3d}, combine spatial hierarchies with LLM parsing and level-by-level retrieval to narrow the search. Their offline construction limits incremental deployment, and early filtering can exclude targets or reference objects needed to resolve coupled constraints. INHerit-SG instead combines structural-event updates with jointly designed memory and retrieval. It maintains the hierarchy incrementally and scores candidates using hierarchical, attribute, and relational evidence, with explicit negation handling (Section~\ref{sec:comparison}).

\subsection{Embodied Memory and Structure-Aware Retrieval}

Embodied memory extends retrieval-augmented generation to spatial and multimodal observations for embodied reasoning and manipulation \cite{lewis2020rag,booker2024embodiedrag,xie2025embodiedrag,anwar2024remembr,yuan2026star,korekata2025affordancerag}. Graph-based systems support language-based planning and navigation (SayPlan, SGNav) \cite{rana2023sayplan,yin2024sgnav}, scene-grounded task decomposition (ASHiTA) \cite{chang2025ashita}, embodied question answering (GraphEQA, Mind Palace) \cite{saxena2025grapheqa,ginting2025mindpalace}, and language-model query interfaces (Structured Interfaces) \cite{ray2025structured}. Relation-aware grounding demonstrates the value of explicit spatial predicates for language-based object localization \cite{kuhn2026relationalscenegraphsobject}. Similarity-based retrieval narrows the context but loses the logical roles of query phrases. Formal query interfaces such as Structured Interfaces scale to large graphs but require per-query query-language generation and assume an already constructed graph.

 


 

\begin{figure*}[t]
\centering
\safeincludegraphics[width=\linewidth]{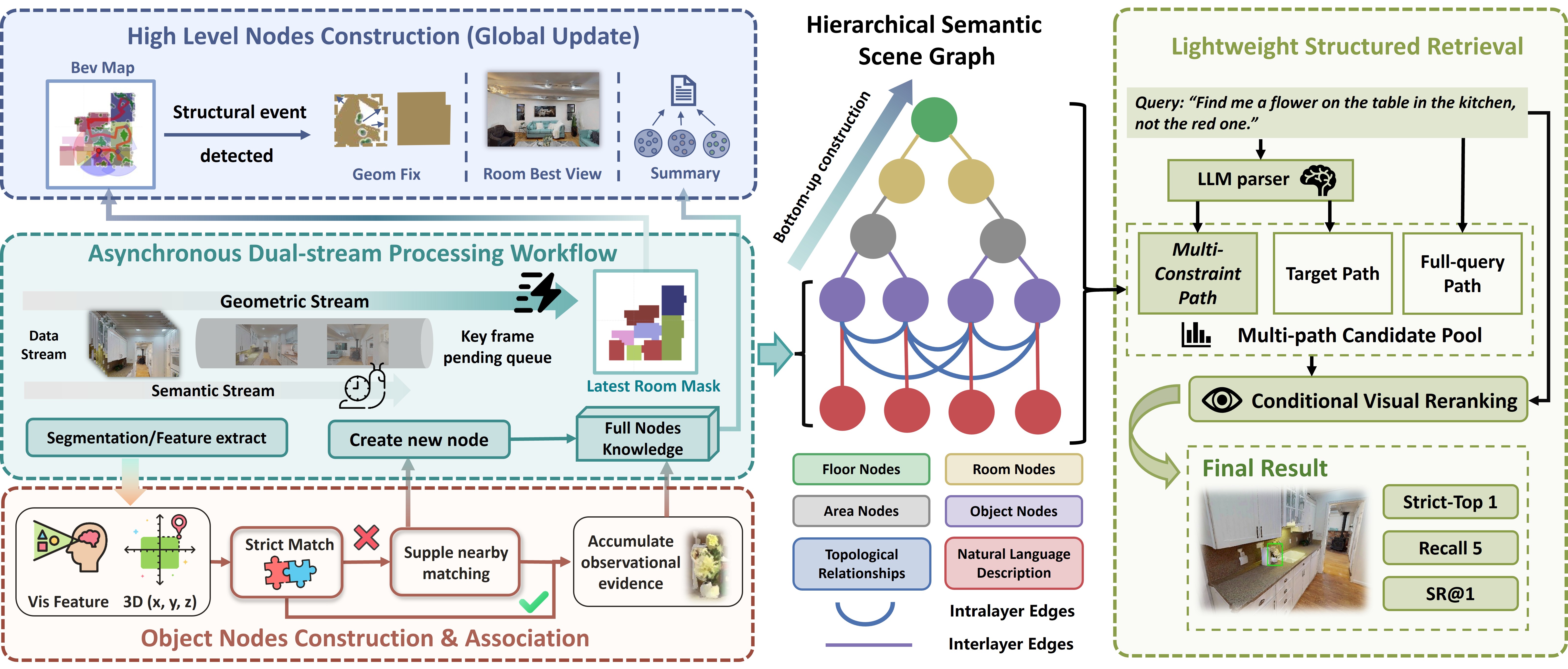}
\caption{\textbf{Overview of incremental hierarchical mapping and structured retrieval in INHerit-SG.}}
\label{fig:framework}
\end{figure*}

\section{Methodology}
 
\subsection{Overview}

INHerit-SG receives a stream of RGB-D observations with camera poses and incrementally constructs a four-level semantic scene graph of the environment. Given a natural language query, it returns an ordered list of candidate object nodes together with the 3D centroid of the top-ranked prediction, which must satisfy the complete instruction. Representation and retrieval are designed jointly, so every constraint type the parser produces maps to evidence in the scene graph (Fig.~\ref{fig:framework}).

\subsection{Dual-Stream Workflow and Object Nodes Construction}

We use floor, room, and object levels common to indoor hierarchical scene graphs \cite{hughes2022hydra,werby2024hovsg} and add functional areas for object groups such as seating or washing areas \cite{vo2019reading}. Floor nodes record vertically separated height ranges and the rooms they contain. Room nodes store semantic labels, 2D masks in the global grid, best-view images, and textual summaries. Area nodes store functional labels and summaries. Object nodes store IDs, semantic labels, 3D centroids and bounding boxes, DINOv3 visual features \cite{simeoni2025dinov3}, VLM-generated descriptions, and references to best-view keyframes. Inter-level edges encode bottom-up membership, forming a floor--room--area--object hierarchy tree. Intra-level spatial relations connect object nodes in a graph.

The mapping pipeline uses a producer--consumer design to separate lightweight geometric tracking from semantic processing of informative keyframes. The geometric stream reads synchronized RGB-D observations and camera poses, detects global height changes to update the floor state, integrates depth observations into a 2D occupancy grid map, and updates geometric room masks. It selects keyframes by comparing DINOv3 visual features of the current frame and the most recent keyframe, treating below-threshold similarity as an approximate indicator of sufficient information gain. The current frame, pose, and geometric context then enter a bounded queue for semantic processing. The geometric stream filters redundant observations and passes keyframes to the semantic worker through the queue ($\approx$ 2 Hz).

The semantic stream consumes accepted keyframes and uses SAM~3 to obtain instance masks \cite{carion2025sam3}. It estimates object positions by back-projecting masked depth measurements into the global coordinate frame. Following common open-vocabulary mapping practice \cite{gu2023conceptgraphs,werby2024hovsg}, a heuristic two-stage greedy strategy associates observations with existing nodes using visual, spatial, and semantic cues. First, candidate matches are globally sorted by cosine similarity between DINOv3 features of observation masks and existing nodes' cropped observations, then matched one-to-one using a high threshold. Second, visual features of unmatched observations are compared with spatially nearby, semantically consistent candidate nodes. Both stages use fixed thresholds.

Matched nodes update their accumulated position information, visual evidence, labels, and best views. Unmatched observations create new nodes, which receive natural language descriptions generated by a VLM from their image crops. The semantic stream then uses 3D positions to construct intra-level spatial relations between objects, including proximity and vertical support relations. These edges are deduplicated and appended to the persistent scene graph. Room and floor transitions drive construction and maintenance of the floor--room--area hierarchy, as described in Section~\ref{sec:maintenance}.

\subsection{Global Update and Higher-Level Node Construction}
\label{sec:maintenance}

Higher-level nodes are maintained through \emph{global updates}, the maintenance passes that construct or refresh pending floor, room, and area nodes from the observations accumulated since the last pass. A global update is triggered when the system detects a room or floor transition, and we refer to these triggers as \emph{structural events}. A floor transition first triggers a global update over all unprocessed higher-level nodes. The system then switches to the current floor's map. Room transitions trigger global updates within a floor.

Updates comprise geometric and semantic processing. Geometric processing cleans up and consolidates room masks on the same floor that the geometric stream has generated ahead of semantic processing. The system removes or merges fragmented masks, regularizes boundaries to right angles, and resolves overlaps to obtain valid rooms.

Semantic processing finalizes higher-level node updates for each valid room in sequence. First, objects are reassigned to rooms using their global positions and the finalized room masks. Second, the system selects the room's best view from the poses at which it has been observed. It chooses the frame with the largest visible room-mask area within the field of view, accounting for wall occlusion. The VLM also scans this image for object expansion, filling in category labels that semantic segmentation misses. Third, spatially nearby member objects are clustered into candidate functional areas. An LLM generates each area's functional label and summary from its member objects' labels and descriptions, and the area node is connected to those objects. Finally, a VLM combines the best-view image with the area summaries to generate the room's semantic label and natural language description. Area nodes are attached to their respective rooms, completing the parent–child membership connections within each room. Section~\ref{sec:efficiency} analyzes the timing of alternative update triggers.

\subsection{Lightweight Structured Retrieval}

The scene graph is serialized into a natural language memory containing the floor, room, area, and object hierarchy, per-object labels and VLM-generated descriptions, textual relation edges, and 3D locations. Functional area nodes remain in the graph to support room-level descriptions. Frame features used for keyframe selection are discarded after use. Object-level DINOv3 features are retained only to associate new observations during later updates. Retrieval operates on the textual memory alone. Best-view images are excluded from it and used only for selective visual reranking. Capable language models can read this memory directly (Section~\ref{sec:model_sensitivity}). The rest of this section develops a lightweight structured interface that avoids full-memory model inference and can also run with a local query parser.

Structured retrieval uses a primary \textit{multi-constraint path} plus two dense paths, the \textit{target path} and the \textit{full-query path}, that increase candidate recall. The multi-constraint path parses query $q$ into typed constraints with explicit polarities:
\begin{equation}
\mathcal{C}(q)=\{c_i=(\tau_i, a_i, r_i, b_i, p_i)\}_{i=1}^{K},
\end{equation}
where $\tau_i$ is the constraint type (target, floor, room, attribute, or relation), $a_i$ the constraint text, $r_i$ and $b_i$ an optional relation predicate and reference constraint, and $p_i\in\{+1,-1\}$ the polarity. Constraints on a reference object remain nested inside the relation constraint.

Each constraint is evaluated at the graph level. Target constraints match object nodes, room constraints match the rooms containing them, and positive floor constraints filter candidates to the target floor before scoring. If no candidate remains, the retriever falls back to a global search. The remaining evidence is combined over the filtered candidate set as a signed weighted sum $S(v \mid q)=\sum_{i=1}^{K} p_i\, w_i\, s_i(v, c_i)$, where $s_i(v,c_i)$ measures semantic similarity or relational compatibility between candidate $v$ and constraint $c_i$, and each weight $w_i$ is assigned and normalized according to the constraints in the query. Negative polarity turns matches with excluded items into penalties, so negated conditions directly affect ranking. Inspired by GraphRAG \cite{edge2024graphrag}, the retriever evaluates relation constraints through explicit graph traversal. For a constraint $c=(a,r,b)$, it traverses candidate $v$'s spatial neighbors, selects edges compatible with predicate $r$, and evaluates reference constraint $b$ on adjacent objects. Nested constraints are resolved inside out.

Relation edges are less reliable when they act as penalties on negated references or must be matched along chains for nested constraints. The multi-constraint path therefore applies \textit{evidence routing} based on the parser's output fields. Queries with negated reference objects bypass relation scoring, queries with nested constraints and a room cue rely on typed constraints and hierarchical evidence, and all other queries use full relational reasoning. Section~\ref{sec:retrieval_analysis} ablates evidence routing separately.

The \textit{target path} and \textit{full-query path} are dense retrieval paths that recover candidates the primary path may miss, matching node text composed of object labels and descriptions. The target path encodes only the target category and non-negated attributes, while the full-query path encodes the raw instruction to retain cues that parsing may discard.

The three ranked lists are fused into the multi-path candidate pool. Within each path, candidates are scored by reciprocal rank, so higher ranks score higher, and the multi-constraint path carries the highest weight. Scores are summed across paths, and the fused ranking fills the pool up to a fixed budget. The multi-constraint path's top prediction is always kept as the first candidate, so the pool widens coverage without overriding the structured ranking. Only when that path returns no candidates does the pool's first candidate become the prediction.

For queries with positive visual attributes, no reference objects, and stored observations available (about one in six queries), the system applies \textit{selective visual reranking}. A VLM jointly compares the candidates' observation images, and its selected candidate moves to first place only if the confidence reaches the specified threshold. Table~\ref{tab:retrieval_ablation} analyzes the retrieval components.

\begin{figure}[t]
\centering
\safeincludegraphics[width=\linewidth]{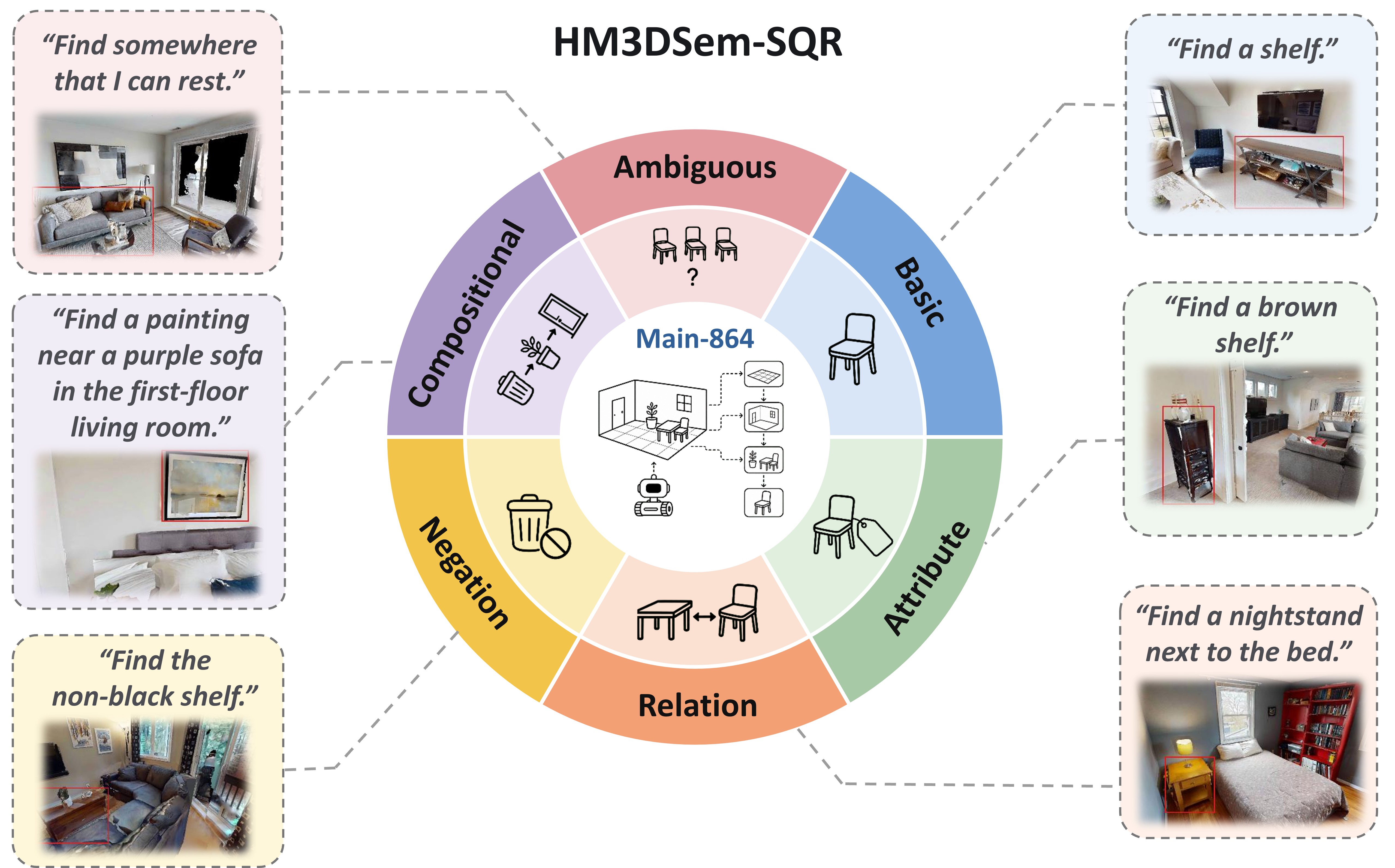}
\caption{\textbf{Query distribution in HM3DSem-SQR Main-864.}}
\label{fig:query_distribution}
\end{figure}

\begin{table*}[t]
\centering
\caption{\textbf{HM3DSem-SQR Main-864 retrieval results (\%).}}
\label{tab:main}
\resizebox{\textwidth}{!}{%
\begin{tabular}{lcccccc|ccc}
\toprule
\rowcolor{tableheader}
\tablehead{Method} & \tablehead{Basic} & \tablehead{Attribute} & \tablehead{Relation} & \tablehead{Negation} & \tablehead{Compositional} & \tablehead{Ambiguous} & \tablehead{Strict Top-1} & \tablehead{Recall@5} & \tablehead{SR@1m} \\
\midrule
ConceptGraphs~\cite{gu2023conceptgraphs}
& 18.06 & 26.39 & 15.28 & 13.89 & 10.42 & 19.44 & 17.25 & 48.61 & 27.08 \\
HOV-SG~\cite{werby2024hovsg}
& 25.69 & 32.64 & 19.44 & 24.31 & 19.44 & 28.47 & 25.00 / 24.54 & 45.25 / 44.91 & 31.60 / 31.02 \\
DualMap$^{*}$~\cite{jiang2025dualmap}
& 32.64 & 36.81 & 22.92 & 22.92 & 17.36 & 23.61 & 26.04 & 54.17 & 32.75 \\
\rowcolor{tableadapt}
Embodied-RAG$^{*}$~\cite{xie2025embodiedrag}
& \textbf{34.03} & 29.17 & 23.61 & 22.92 & 24.31 & 19.44 & 25.58 / 25.69 & 38.77 / 42.36 & 29.63 / 31.83 \\
OVSG~\cite{pmlr-v229-chang23b}
& 25.00 & 25.00 & \textbf{29.86} & 27.08 & 21.53 & 28.47 & 26.16 / 25.58 & 54.98 / 55.09 & 32.18 / 31.60 \\
ReasoningGraph~\cite{puigjaner2026relationshipawarehierarchical3dscene}$^{*}$
& 17.36 & 37.50 & 18.06 & 11.81 & 11.81 & 20.14 & 19.44 / 19.79 & 34.26 / 34.84 & 22.92 / 23.03 \\
KeySG$^{*}$~\cite{werby2026keysghierarchicalkeyframebased3d}
& 29.17 & 25.00 & 29.17 & 18.75 & 28.47 & \textbf{29.17} & 26.62 / 22.92 & 39.58 / 32.99 & 32.64 / 27.43 \\
\midrule
\rowcolor{tableours}
\textbf{INHerit-SG (Ours)}
& 32.64 & \textbf{45.14} & \textbf{29.86} & \textbf{27.78} & \textbf{34.72} & 27.78 & \textbf{32.99} & \textbf{62.27} & \textbf{37.04} \\
\bottomrule
\end{tabular}}
\par\vspace{4pt}\noindent\parbox{\textwidth}{\scriptsize $^{*}$ denotes evaluator-only adaptation. The yellow row uses the INHerit-SG mapping front-end. Per-category columns report Strict Top-1 under each method's official setting. In the overall columns, A / B reports the official / unified setting (Sections~\ref{sec:setup} and~\ref{sec:comparison}).}
\end{table*}

\section{HM3DSem-SQR Benchmark}
\label{sec:benchmark}

We construct HM3DSem-SQR from all 36 validation scenes in HM3D-Sem \cite{ramakrishnan2021hm3d}. A human operator records one exploration trajectory per scene with synchronized RGB-D observations and camera poses. Habitat-Sim re-renders each trajectory to obtain semantic instance observations. Fusing these with the HM3D-Sem scene annotations provides object identities, categories, room membership, floor assignments, and geometric extents.

GPT-4o generates a short free-text description for each visible object. Ground-truth spatial relations between objects come from geometric tests on oriented 3D bounding boxes. Semantic refinement and filtering then keep four relation types: \emph{near}, \emph{on}, \emph{under}, and \emph{inside}.

Each query is generated from the scene records and paired with a hard negative that violates only a designated constraint. The full release contains 6,084 automatically generated instructions across 14 fine-grained query types. All reported HM3DSem-SQR results use manually reviewed subsets with valid categories, nonempty target sets, hard negatives, and keyframe evidence of the targets. Queries fall into six mutually exclusive categories: \textbf{Basic}, \textbf{Attribute}, \textbf{Relation}, \textbf{Negation}, \textbf{Compositional}, and \textbf{Ambiguous}. Compositional queries require the target to satisfy multiple interrelated constraints, or the reference object itself to satisfy its own constraints.

We construct \textbf{Main-864} with 24 queries from each of the 36 scenes and 144 queries per category, balanced across fine-grained query types. All 864 queries are manually reviewed. \textbf{Core-432} is selected from Main-864 with 12 queries per scene and the same category distribution. \textbf{MapAblation-96} further selects 8 scenes of different sizes with 12 queries per scene, 96 in total.

\section{Experiments}

\subsection{Experimental Setup}
\label{sec:setup}
\paragraph{Datasets and Evaluation Subsets.}
We use the fixed splits in Section~\ref{sec:benchmark}: Main-864 for system-level comparisons across 36 validation scenes, Core-432 for controlled retrieval analysis, and MapAblation-96 for foundation model sensitivity and mapping backend ablations.
 

\paragraph{Metrics.}
The primary metric, \textbf{strict Top-1 accuracy} (\%), counts a query as successful when the top-ranked prediction corresponds to a ground-truth object whose instance ID belongs to the query's target set, which may contain multiple valid instances. Strict Top-1 evaluation uses a strict geometric correspondence between predictions and ground truth. A predicted node matches a simulator object if its 3D center lies inside the object's oriented 3D bounding box or within 0.25 m of the box. When several adjacent or overlapping boxes match and their distances differ by no more than 0.05 m, the evaluator selects a unique instance by category. The selection follows a predefined correspondence table between predicted labels and simulator labels, reviewed by hand. Recall@5 uses the same alignment procedure to check whether a correct target appears among the top five candidates. SR@1m checks only whether the predicted position lies within 1\,m of any correct target. We report results separately for all six query categories.

\paragraph{Implementation and Fixed Settings.}
Unless explicitly varied, trajectories, queries, evaluators, prompts, thresholds, and random seeds remain fixed. During mapping, INHerit-SG uses Qwen3-VL-Flash for object descriptions, object expansion, and room summaries, and Qwen3.6-Flash for semantic aggregation in higher-level node construction. At query time, Qwen3.6-Flash parses instructions and the structured retriever ranks candidates. Gemini~2.5~Flash selectively reranks eligible attribute queries. This default configuration uses hosted models. The local-model configurations in Section~\ref{sec:model_sensitivity} separately evaluate retrieval without hosted query-time inference.

\paragraph{Compared Methods.}
We compare against ConceptGraphs \cite{gu2023conceptgraphs}, HOV-SG \cite{werby2024hovsg}, DualMap \cite{jiang2025dualmap}, Embodied-RAG \cite{xie2025embodiedrag}, OVSG \cite{pmlr-v229-chang23b}, ReasoningGraph \cite{puigjaner2026relationshipawarehierarchical3dscene}, and KeySG \cite{werby2026keysghierarchicalkeyframebased3d}. All methods build maps from the same RGB-D trajectories and poses and are evaluated on the same Main-864 queries under a common object alignment protocol. Embodied-RAG is the only exception, since it lacks an RGB-D mapping front-end and builds its semantic forest on ours. Adaptations pass only query text through each method's own retrieval mechanism, without ground-truth target IDs, positions, or pre-parsed constraints. Each baseline runs under two protocols that differ only in the foundation models. The first is its official configuration, which keeps the generative models released with its official implementation. The second is a unified setting, which uses the same models as INHerit-SG. Qwen3.6-Flash handles text calls, and Qwen3-VL-Flash and Gemini~2.5~Flash handle visual calls in mapping and retrieval. The unified setting separates design from model capability. Non-generative components and ReasoningGraph's map-bound DeepSeek-VL2 inspector remain unchanged. Implementation details and adapters will be released with the code.


\begin{table*}[t]
\centering
\caption{\textbf{Structured retrieval ablation on Core-432 (\%).}}
\label{tab:retrieval_ablation}
\resizebox{\textwidth}{!}{%
\begin{tabular}{lccccccc|cc}
\toprule
\rowcolor{tableheader}
\tablehead{Retrieval strategy} & \tablehead{Basic} & \tablehead{Attribute} & \tablehead{Relation} & \tablehead{Negation} & \tablehead{Compositional} & \tablehead{Ambiguous} & \tablehead{Strict Top-1} & \tablehead{Recall@5} & \tablehead{Mean local ms/q} \\
\midrule
Typed constraint retrieval
& 27.78 & 26.39 & 20.83 & 23.61 & 23.61 & 20.83 & 23.84 & 52.78 & 50.8 \\
+ Hierarchy-aware retrieval
& 31.94 & 31.94 & 19.44 & 27.78 & 18.06 & 27.78 & 26.16 & 55.56 & 78.0 \\
+ Relation graph reasoning
& 31.94 & 31.94 & 27.78 & 23.61 & 16.67 & 30.56 & 27.08 & 56.25 & 49.5 \\
+ Evidence routing
& 31.94 & 31.94 & 27.78 & 27.78 & 25.00 & 31.94 & 29.40 & 57.64 & 36.9 \\
+ Multi-path candidate pool
& 31.94 & 31.94 & 27.78 & 27.78 & 27.78 & 31.94 & 29.86 & 62.73 & 263.0 \\
+ Selective visual reranking
& 31.94 & 45.83 & 27.78 & 27.78 & 27.78 & 33.33 & 32.41 & 62.73 & 263.0 + 1,227$^{\dagger}$ \\
\bottomrule
\end{tabular}}
\par\vspace{4pt}\noindent\parbox{\textwidth}{\scriptsize $^{\dagger}$ VLM reranking (Gemini 2.5 Flash) runs on $\sim1/6$ of queries; measured 135 calls, 17m40s total ($\approx$ 7.9 s per eligible query) on Main-864. Latency excludes query parsing.}
\end{table*}

\subsection{Comparison with Existing Systems}
\label{sec:comparison}

On Main-864, INHerit-SG achieves 32.99\% strict Top-1, 62.27\% Recall@5, and 37.04\% SR@1m (Table~\ref{tab:main}). Under the original model configurations, it exceeds the strongest baseline KeySG by 6.37 percentage points in strict Top-1. Under the unified model setting, HOV-SG, Embodied-RAG, OVSG, and ReasoningGraph change by at most 0.6 points, whereas KeySG decreases from 26.62\% to 22.92\%. INHerit-SG leads KeySG by 10.07 points and DualMap, the strongest baseline in this setting, by 6.94 points. The advantage persists after model matching, which supports the contribution of the representation and retrieval design.

INHerit-SG is the best or tied for best on four of the six query categories, with the largest gains over the best baseline on attribute (7.6 points) and compositional (6.3 points) queries. It trails only on basic and ambiguous queries, by at most two queries each. Ambiguous targets lack decisive constraints and depend more on commonsense priors than on structural scene evidence. The baselines' profiles reflect their designs. KeySG nearly matches INHerit-SG on relation queries (29.17\% vs. 29.86\%), leads on ambiguous ones (29.17\% vs. 27.78\%), and is the strongest baseline on compositional ones (28.47\% vs. 34.72\%), the categories its LLM-based query decomposition targets. It trails by 20.14 and 9.03 points on attribute and negation queries, and its Recall@5 reaches only 39.58\% (vs. 62.27\%). Early hierarchical filtering is one possible contributor to this coverage gap (Section~\ref{RW:B}). OVSG and DualMap show the opposite limitation. Dense similarity matching gives them the highest baseline Recall@5 (54.98\% and 54.17\%), yet both trail by over six points on strict Top-1, showing that candidate coverage alone does not ensure correct final selection. INHerit-SG is the only method that exceeds 60\% Recall@5 and 30\% strict Top-1.

\subsection{Failure Attribution and Diagnosis}

\begin{figure}[t]
\centering
\safeincludegraphics[width=.94\columnwidth]{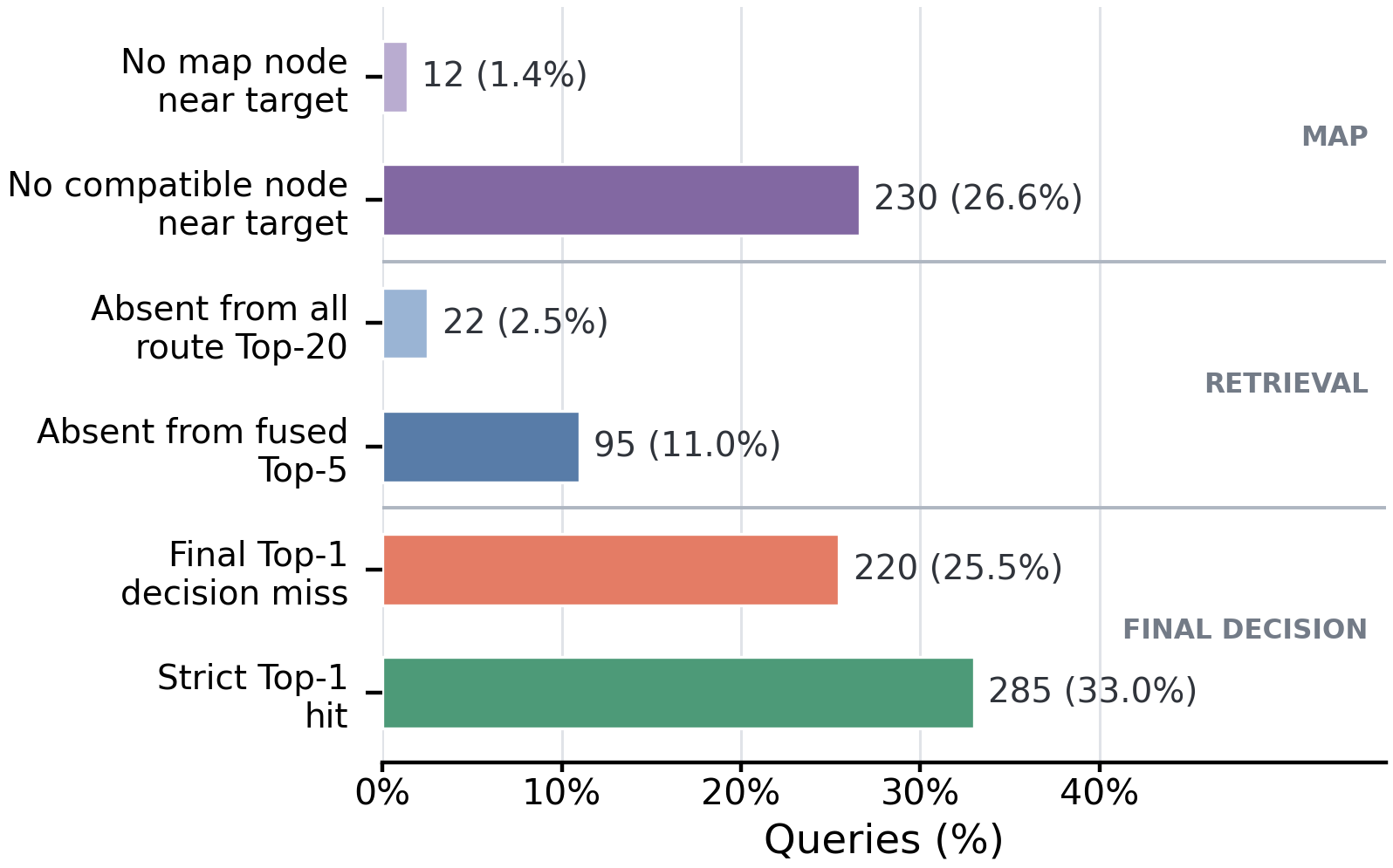}
\caption{\textbf{Component-level attribution for INHerit-SG on Main-864.}}
\label{fig:inherit_failure}
\end{figure}

Saved maps and retrieval records support a stage-wise diagnosis of failed queries (Fig.~\ref{fig:inherit_failure}). Each failure is assigned to the first unmet diagnostic check. Two categories dominate. 230 queries (26.6\%) have geometrically nearby map nodes but none with a category compatible with the target, and 220 queries (25.5\%) pass the preceding checks and retain a correct target in the fused candidate list but fail at final Top-1 selection. Only 12 queries lack any map node near the target. These diagnostic checks differ from the final identity-alignment metric. The map-semantic check requires category compatibility, whereas the evaluator uses it to resolve geometric ambiguity and permits geometric fallback. Intermediate and final candidate lists can also differ. The stage counts therefore do not directly decompose Recall@5. The diagnosis motivates richer multi-view semantic integration and more reliable candidate verification.

\subsection{External Generalization on ViGiL3D}
We evaluate on ViGiL3D \cite{wang2025vigil3dlinguisticallydiversedataset}, which contains 349 queries across 26 ScanNet and 10 ScanNet++ scenes, using its official release. The GT-box oracle evaluation covers all 349 queries, and the predicted-box evaluation uses Mask3D ScanNet200 proposals for the 239 ScanNet queries. Table~\ref{tab:vigil3d} reports the ScanNet subset under both protocols, with baseline results from the ViGiL3D paper. To handle multiple-target and no-target queries, Gemini 2.5 Flash selects an object set or an empty set from retrieved candidates using the query, scene memory, and candidate views, without access to ground-truth target identities or counts. We report box-matching F1 and Exact-set Accuracy, which requires recovering all targets without extra predictions.

Without task-specific training, INHerit-SG achieves 20.84\% and 18.33\% F1 at IoU 0.25 and 0.50, outperforming all open-vocabulary zero-shot baselines and the strongest trained method, 3D-VisTA (15.70\% and 13.20\%). The GT-box oracle removes proposal localization errors and gives a complementary comparison. INHerit-SG achieves 27.20\% Exact-set Accuracy and 29.11\% F1, compared with 26.20\% and 26.80\% for PQ3D. Together, the two protocols indicate that the retrieval design transfers to independently authored queries and different geometric inputs.

\subsection{Retrieval Analysis}
\label{sec:retrieval_analysis}

The results above evaluate the complete pipeline. We next isolate where the retrieval gains come from. All configurations in Table~\ref{tab:retrieval_ablation} share identical INHerit-SG scene graphs and Core-432 queries, and they differ only in which query-time components are active, with components added cumulatively.

\begin{table}[t]
\centering
\caption{\textbf{ViGiL3D ScanNet results (\%).}}
\label{tab:vigil3d}
\footnotesize
\setlength{\tabcolsep}{0pt}
\begin{tabular}{@{}>{\hspace*{3pt}\raggedright\arraybackslash}p{.60\columnwidth}>{\centering\arraybackslash}p{.20\columnwidth}>{\centering\arraybackslash}p{.20\columnwidth}@{}}
\toprule
\rowcolor{tablecontrol}
\multicolumn{3}{l}{\hspace*{3pt}\textbf{(a) ScanNet predicted boxes (239 queries)}} \\
\rowcolor{tableheader}
\tablehead{Method} & \tablehead{F1@0.25} & \tablehead{F1@0.50} \\
\midrule
\rowcolor{tablecontrol}
\multicolumn{3}{l}{\hspace*{3pt}\emph{Open-vocabulary and zero-shot baselines}} \\
OpenScene & 1.70 & 1.20 \\
LERF & 2.10 & 2.10 \\
ZSVG3D & 6.70 & 5.80 \\
LLM-Grounder & 5.30 & 3.10 \\
\rowcolor{tablecontrol}
\multicolumn{3}{l}{\hspace*{3pt}\emph{Trained baselines}} \\
3D-VisTA & 15.70 & 13.20 \\
3D-GRAND & 15.30 & 11.80 \\
PQ3D & 5.60 & 5.10 \\
\rowcolor{tableours}
\textbf{INHerit-SG (Ours, zero-shot)} & \textbf{20.84} & \textbf{18.33} \\
\midrule
\rowcolor{tablecontrol}
\multicolumn{3}{l}{\hspace*{3pt}\textbf{(b) ScanNet GT-box oracle (239 queries)}} \\
\rowcolor{tableheader}
\tablehead{Method} & \tablehead{Exact-set Acc.} & \tablehead{F1} \\
\midrule
ZSVG3D & 18.90 & 12.20 \\
PQ3D & 26.20 & 26.80 \\
\rowcolor{tableours}
\textbf{INHerit-SG (Ours)} & \textbf{27.20} & \textbf{29.11} \\
\bottomrule
\end{tabular}
\end{table}

\begin{table}[t]
\centering
\caption{\textbf{Foundation model sensitivity on MapAblation-96.}}
\label{tab:backend}
\label{tab:construction}
\footnotesize
\setlength{\tabcolsep}{0pt}
\textbf{(a) Retrieval interfaces}\par\vspace{2pt}
\begin{tabular}{@{}>{\hspace*{3pt}\raggedright\arraybackslash}p{.26\columnwidth}>{\raggedright\arraybackslash}p{.16\columnwidth}>{\centering\arraybackslash}p{.13\columnwidth}>{\centering\arraybackslash}p{.17\columnwidth}>{\centering\arraybackslash}p{.15\columnwidth}>{\centering\arraybackslash}p{.13\columnwidth}@{}}
\toprule
\rowcolor{tableheader}
\tablehead{Model} & \tablehead{Interface} & \tablehead{Top-1} & \tablehead{Recall@5} & \tablehead{API Cost} & \tablehead{Time} \\
\midrule
\multirow{2}{*}{Gemini 2.5 Pro}
    & Direct     & 39.58 & 68.75 & 0.0887 & 17.34 \\
    & Structured & 31.25 & 50.00 & 0.0140 & 10.70 \\
\cmidrule{1-6}
\multirow{2}{*}{Qwen3.6-Flash}
    & Direct     & 36.46 & 65.63 & 0.0084 & 4.00 \\
    & Structured & 30.21 & 54.17 & 0.0005 & 2.75 \\
\cmidrule{1-6}
\multirow{2}{*}{Qwen2-VL-7B}
    & Direct     & 14.58 & 20.83 & 0      & 10.03 \\
    & Structured & 27.08 & 54.17 & 0      & 3.08 \\
\cmidrule{1-6}
\multirow{2}{*}{Qwen2.5-VL-3B}
    & Direct     & 5.21  & 15.63 & 0      & 4.17 \\
    & Structured & 27.08 & 57.29 & 0      & 3.11 \\
\bottomrule
\end{tabular}
\par\vspace{2pt}\noindent\parbox{\columnwidth}{\scriptsize Top-1 and Recall@5 in \%; cost in USD per query; time in seconds per query.}
\par\vspace{6pt}
\textbf{(b) Construction backends (\%)}\par\vspace{2pt}
\begin{tabular}{@{}>{\hspace*{3pt}\raggedright\arraybackslash}p{.52\columnwidth}>{\centering\arraybackslash}p{.19\columnwidth}>{\centering\arraybackslash}p{.15\columnwidth}>{\centering\arraybackslash}p{.14\columnwidth}@{}}
\toprule
\rowcolor{tableheader}
\tablehead{Visual + aggregation model} & \tablehead{Top-1} & \tablehead{Recall@5} & \tablehead{SR@1m} \\
\midrule
Gemini 2.5 Flash + Qwen3.6-Flash & \textbf{33.33} & 53.13 & 37.50 \\
GPT-4o + Qwen3.6-Flash & \textbf{33.33} & 54.17 & 38.54 \\
Qwen3-VL-Flash + Gemini 2.5 Flash & 32.29 & 51.04 & \textbf{39.58} \\
Qwen3-VL-Flash + GPT-4o & 32.29 & 53.13 & 38.54 \\
\rowcolor{tableours}
Qwen3-VL-Flash + Qwen3.6-Flash$^{\dagger}$ & 32.29 & \textbf{56.25} & 38.54 \\
\bottomrule
\end{tabular}
\par\vspace{2pt}\noindent\parbox{\columnwidth}{\scriptsize $^{\dagger}$ denotes the configuration used by INHerit-SG.}
\end{table}

The components contribute through distinct mechanisms. Hierarchical evidence restricts the search scope and raises overall accuracy from 23.84\% to 26.16\%. Relation reasoning helps relation queries most, but it is not uniformly reliable. Where edges act as penalties (negation) or must be matched along chains (compositional), residual edge errors are amplified and both categories degrade. Evidence routing addresses this query dependence. It restores negation and compositional accuracy (to 27.78\% and 25.00\%), raising overall strict Top-1 to 29.40\% (scene-level paired sign-flip permutation test, two-sided, $p = 0.012$).

Recall and accuracy are then improved separately. The multi-path candidate pool brings more correct targets into the candidate list (Recall@5 57.64\% to 62.73\%), at an additional 226.1 ms of local latency per query. It can be disabled for latency-sensitive deployment. Selective visual reranking converts this recall into accuracy, reaching 32.41\% overall ($p=0.023$) with attribute queries improving most (31.94\% to 45.83\%). The full pipeline improves over typed-constraint retrieval alone by 8.57 points, and each component accounts for a distinct part of this gain.

\subsection{Foundation Model Sensitivity}
\label{sec:model_sensitivity}

We compare two retrieval interfaces on the same MapAblation-96 queries and serialized INHerit-SG memory (Table~\ref{tab:backend}). \textit{Direct} reads the complete textual memory ($\approx$33k tokens per scene) and returns up to five object IDs ranked by relevance. \textit{Structured} uses the listed model to parse constraints, then applies fixed local retrieval without evidence routing or visual reranking. The Main-864 system comparison uses the full retrieval pipeline. Component ablations and external evaluation follow the configurations described in their respective subsections.

For powerful hosted models, \textit{Direct} serves as a reference for achievable accuracy (39.58\% vs.\ 31.25\% strict Top-1 for Gemini 2.5 Pro), but at 6.3--16.1$\times$ the per-query API cost. Smaller local models show the opposite pattern. \textit{Structured} raises strict Top-1 from 14.58\% to 27.08\% for Qwen2-VL-7B and from 5.21\% to 27.08\% for Qwen2.5-VL-3B. These results suggest that fitting scene memory within a model's context window does not ensure direct retrieval, and structured access can help smaller models resolve the query. Across the five construction-backend combinations, strict Top-1 varies by at most one query on this subset (Table~\ref{tab:construction}). The same memory thus supports direct reading by strong models and structured access by smaller models.

\subsection{Efficiency and Maintenance}
\label{sec:efficiency}

Table~\ref{tab:maintenance} compares structural-event updates (global updates only at room/floor transitions) against fixed-period updates every $N$ accepted keyframes on MapAblation-96, with all other components shared. Update counts and times are per-scene averages covering global updates only. Event triggering matches the accuracy of the sparsest fixed period ($N{=}18$) and trails the most frequent one ($N{=}6$) by only 1.04 points, while incurring the lowest total update time (394.09 s vs. 441.40 to 804.51 s). Structural events are thus an effective and cheaper trigger for expensive global updates.

\begin{table}[t]
\centering
\caption{\textbf{Maintenance-policy ablation on MapAblation-96.}}
\label{tab:maintenance}
\footnotesize
\setlength{\tabcolsep}{0pt}
\begin{tabular}{@{}>{\hspace*{3pt}\raggedright\arraybackslash}p{.40\columnwidth}>{\centering\arraybackslash}p{.18\columnwidth}>{\centering\arraybackslash}p{.21\columnwidth}>{\centering\arraybackslash}p{.21\columnwidth}@{}}
\toprule
\rowcolor{tableheader}
\tablehead{Policy} & \tablehead{Top-1 (\%)} & \tablehead{Avg.\ Updates} & \tablehead{Avg.\ Time (s)} \\
\midrule
Fixed-period, N=6 & 33.33 & 64.62 & 804.51 \\
Fixed-period, N=9 & 32.29 & 42.88 & 654.18 \\
Fixed-period, N=18 & 32.29 & 21.25 & 441.40 \\
\rowcolor{tableours}
Structural events & 32.29 & 23.62 & 394.09 \\
\bottomrule
\end{tabular}
\par\vspace{2pt}\noindent\parbox{\columnwidth}{\scriptsize Update counts and times are per-scene averages covering global updates only.}
\end{table}


\begin{table}[t]
\centering
\caption{\textbf{Real-robot Top-1 success after human review (\%).}}
\label{tab:realworld}
\footnotesize
\setlength{\tabcolsep}{0pt}
\begin{tabular}{>{\hspace*{3pt}\raggedright\arraybackslash}p{.34\columnwidth}>{\centering\arraybackslash}p{.22\columnwidth}>{\centering\arraybackslash}p{.22\columnwidth}>{\centering\arraybackslash}p{.22\columnwidth}}
\toprule
\rowcolor{tableheader}
\tablehead{Method} & \tablehead{Simple} & \tablehead{Complex} & \tablehead{Overall} \\
\midrule
HOV-SG~\cite{werby2024hovsg} & 45.83 & 32.35 & 40.24 \\
DualMap$^{*}$~\cite{jiang2025dualmap} & 18.75 & 20.59 & 19.51 \\
OVSG~\cite{pmlr-v229-chang23b} & \textbf{54.17} & 26.47 & 42.68 \\
KeySG$^{*}$~\cite{werby2026keysghierarchicalkeyframebased3d} & 37.50 & 44.12 & 40.24 \\
\rowcolor{tableours}
\textbf{INHerit-SG (Ours)} & \textbf{54.17} & \textbf{50.00} & \textbf{52.44} \\
\bottomrule
\end{tabular}
\par\vspace{4pt}\noindent\parbox{\textwidth}{\scriptsize $^{*}$ denotes adapted mapping or evaluation configurations.}
\end{table}

\begin{figure}[!t]
\centering
\safeincludegraphics[width=\linewidth]{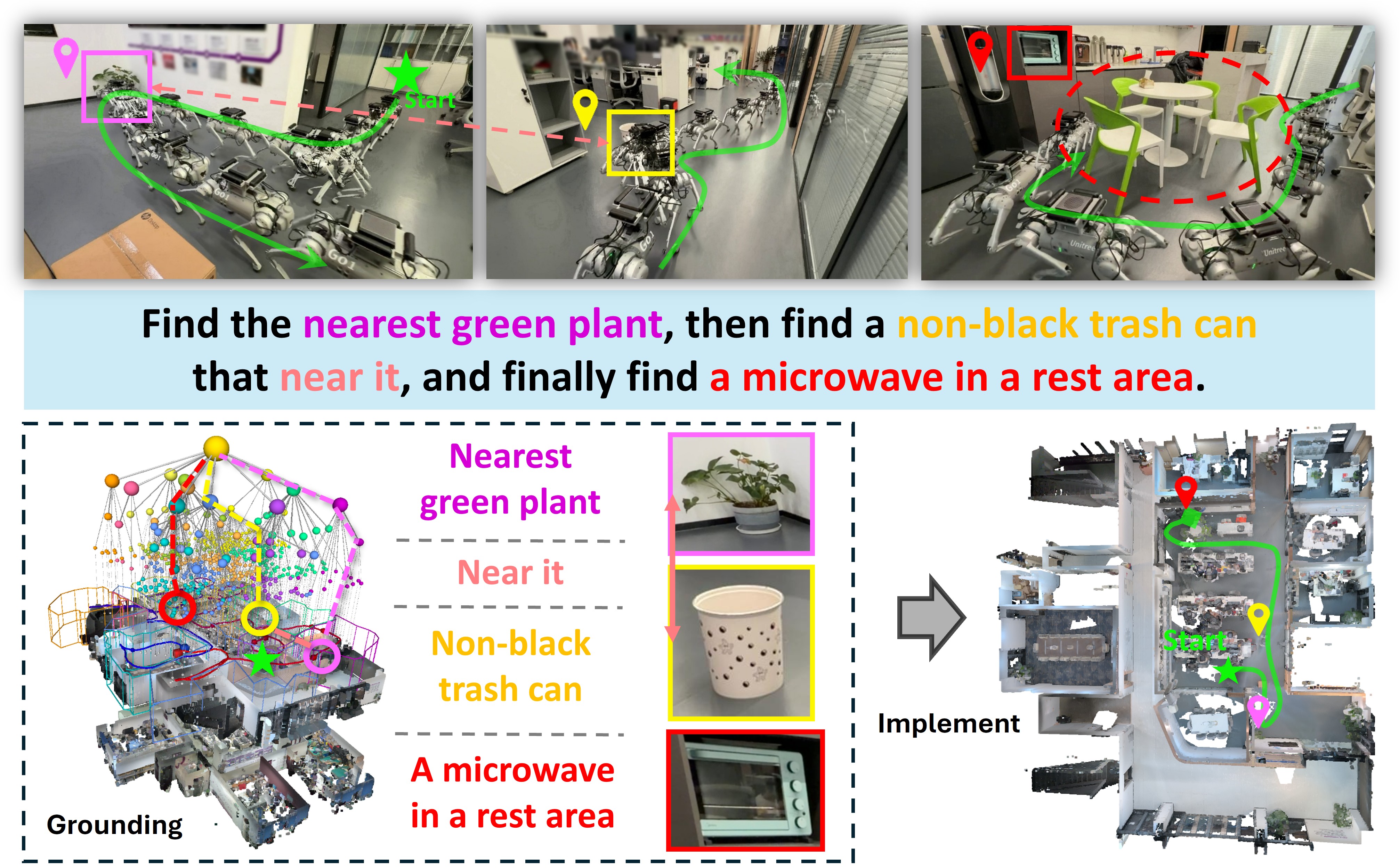}
\caption{\textbf{Qualitative real-world demonstration showing the retrieved target sequence and robot trajectory.}}
\label{fig:real}
\end{figure}

\subsection{Real-World Evaluation}

We evaluate the complete system in three real indoor environments: two 200 m$^2$ office/laboratory spaces and a 400 m$^2$ two-floor teaching and research space, with 82 queries in total. All methods build maps from the same recorded RGB-D trajectories and run on the same RTX 4090 workstation. Queries involving explicit negation, reference-object relations, coupled referring expressions, or unnamed targets are labeled complex (34), including ambiguous queries whose targets lack decisive constraints. The rest are labeled simple (48), with labels fixed before aggregation. A committee of three VLMs (GPT-5.5, Gemini 3.1 Pro, Claude Opus 4.8) judges each prediction against the query and its visual evidence, voting supported, uncertain, or unsupported. The results are then verified by human review, and we report the human-verified success rates.

INHerit-SG achieves the highest overall success rate at 52.44\%, leading the strongest baseline (OVSG, 42.68\%) by 9.76 points (Table~\ref{tab:realworld}). It ties OVSG on simple queries (54.17\%) and leads all methods on complex ones (50.00\%). KeySG remains the closest competitor on complex queries (44.12\%) but trails on simple ones (37.50\% vs. 54.17\%).

\section{Conclusion}

We presented INHerit-SG, an incremental hierarchical semantic scene graph for compositional embodied object retrieval, and constructed the HM3DSem-SQR benchmark to study such retrieval tasks. By combining asynchronous geometric and semantic processing, structural event-driven hierarchy maintenance, and lightweight structured retrieval, the system organizes continuously acquired observations into queryable, extensible spatial memory and provides downstream reasoning modules with hierarchical and relational evidence at different granularities.

\bibliographystyle{IEEEtran}
\bibliography{inherit_aaai27}

\end{document}